\documentclass[letterpaper]{article} 
\usepackage{aaai2026}  
\usepackage{amsmath}
\usepackage{times}  
\usepackage{helvet}  
\usepackage{courier}  
\usepackage[hyphens]{url}  
\usepackage{graphicx} 
\usepackage{natbib}  
\usepackage{caption} 
\usepackage{algorithm}
\usepackage{algorithmic}
\usepackage[svgnames, table, xcdraw]{xcolor}
\usepackage{tikz}
\usepackage[most]{tcolorbox}
\usepackage{xparse} 
\usepackage{lipsum}
\usepackage{fontawesome5}
\newcounter{boxnumber}
\tcbset{
  myfancytitle/.style={
    colback=cyan!3,
    colframe=cyan!50!black,
    coltitle=black,
    fonttitle=\bfseries\large,
    center title,
    enhanced,
    attach boxed title to top center={yshift=-2mm},
    boxed title style={
      colback=cyan!10,
      colframe=cyan!70!black,
      boxrule=0.8pt,
      arc=12pt,
      drop shadow,
      left=6pt,
      right=6pt,
      top=4pt,
      bottom=4pt,
    }
  },
  numberedboxstyle/.style={
    enhanced,
    width=\textwidth,
    colback=cyan!1,
    colframe=cyan!60!black,
    coltitle=black,
    fonttitle=\bfseries,
    breakable,
    before skip=6pt,
    after skip=6pt,
    colbacktitle=cyan!10, 
    coltitle=black,       
    center title,         
    boxed title style={
      enhanced,
      arc=10pt,
      boxrule=0.7pt,
      left=6pt,
      right=6pt,
      top=4pt,
      bottom=4pt,
      drop shadow
    }
    center title
  }
}

\NewDocumentCommand{\numberedbox}{O{} +m}{%
  \refstepcounter{boxnumber}%
  \IfNoValueTF {#1}
    {
      \begin{tcolorbox}[numberedboxstyle]
        #2
      \end{tcolorbox}
    }
    {
       \begin{tcolorbox}[
        numberedboxstyle,
        title={#1}
      ]
        #2
      \end{tcolorbox} 
    }
}
\usepackage{newfloat}
\usepackage{listings}
\DeclareCaptionStyle{ruled}{labelfont=normalfont,labelsep=colon,strut=off} 
\floatstyle{ruled}
\newfloat{listing}{tb}{lst}{}
\floatname{listing}{Listing}
\nocopyright 

\title{Dynamic Knowledge Exchange and Dual-diversity Review: Concisely Unleashing the Potential of a Multi-Agent Research Team}

\author {
    Weilun Yu\textsuperscript{\rm 1,\rm 2},
    Shixiang Tang\textsuperscript{\rm 3},
    Yonggui Huang\textsuperscript{\rm 4},
    Nanqing Dong\textsuperscript{\rm 3},
    Li Fan\textsuperscript{\rm 1,\rm 2},
    Honggang Qi\textsuperscript{\rm 5},
    Wei Liu\textsuperscript{\rm 6},
    Xiaoli Diao\textsuperscript{\rm 7},
    Xi Chen\textsuperscript{\rm 2}\thanks{Corresponding author},
    Wanli Ouyang\textsuperscript{\rm 3}
}
\affiliations {
    \textsuperscript{\rm 1}School of Geographic Sciences, East China Normal University\\
    \textsuperscript{\rm 2}Software Engineering Institute, East China Normal University\\
    \textsuperscript{\rm 3}Shanghai Artificial Intelligence Laboratory\\
    \textsuperscript{\rm 4}Peking University\\
    \textsuperscript{\rm 5}School of Computer Science and Technology, University of Chinese Academy of Sciences\\
    \textsuperscript{\rm 6}Bytedance Inc\\
    \textsuperscript{\rm 7}Canadian Collaborative Entrepreneurs Organization Inc\\
    xchen@geo.ecnu.edu.cn
}

\usepackage{bibentry}

\begin{document}

\maketitle

\begin{abstract}
Scientific progress increasingly relies on effective collaboration among researchers, a dynamic that large language models (LLMs) have only begun to emulate. While recent LLM-based scientist agents show promise in autonomous scientific discovery, they often lack the interactive reasoning and evaluation mechanisms essential to real-world research.
We propose IDVSCI (Internal Discussion and Vote SCIentists), a multi-agent framework built on LLMs that incorporates two key innovations: a Dynamic Knowledge Exchange mechanism enabling iterative feedback among agents, and a Dual-Diversity Review paradigm that simulates heterogeneous expert evaluation. These components jointly promote deeper reasoning and the generation of more creative and impactful scientific ideas.
To evaluate the effectiveness and generalizability of our approach, we conduct experiments on two datasets: a widely used benchmark in computer science and a new dataset we introduce in the health sciences domain. Results show that IDVSCI consistently achieves the best performance across both datasets, outperforming existing systems such as AI Scientist and VIRSCI. These findings highlight the value of modeling interaction and peer review dynamics in LLM-based autonomous research.
\end{abstract}

\begin{links}
    \link{Code}{https://github.com/vvvlun/IDVSCI}
    \link{Datasets (Computer Sciences)}{https://drive.google.com/drive/folders/1ZwWMBQ5oK-l4VuzMa60GbMND0g2EIxIu?usp=sharing}
    \link{Datasets (Health Sciences)}{https://drive.google.com/file/d/1AedrqqaGgml7ExGGWBGmST413JSZ3jzF/view?usp=drive_link}
\end{links}

\section{Introduction}

With the remarkable progress of artificial intelligence (AI), autonomous scientific discovery has emerged as a promising direction for reshaping how research is conducted. Recent work such as AI Scientist~\cite{lu2024ai} demonstrates the potential of large language models (LLMs) to generate scientific papers in an end-to-end manner, highlighting their capacity to drive automated discovery. However, AI Scientist adopts a single-agent design that falls short in reflecting the inherently collaborative nature of real-world scientific research~\cite{gauch2003scientific}.

\begin{figure}[!t]
    \centering
    \includegraphics[width=\columnwidth]{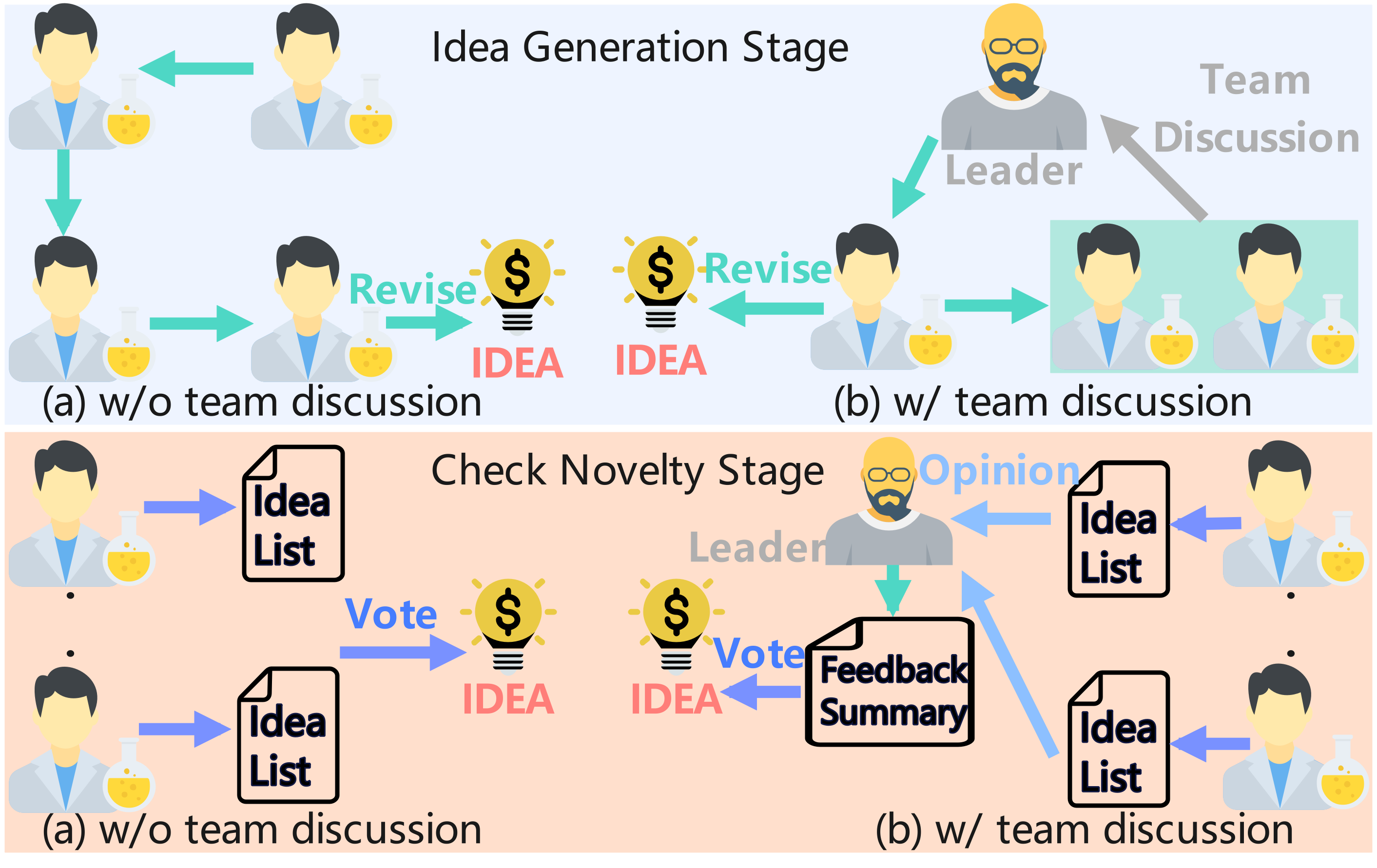}
    \caption{Illustration of the workflow comparison between conventional approaches and our proposed methodology (IDVSCI), showing how agent-based collaboration enhances both the Idea Generation Stage and the Check Novelty Stage.}
    \label{fig_1}
    \end{figure}

To better emulate authentic scientific collaboration, LLM-based multi-agent systems are essential. Traditional multi-agent systems rely on structured protocols and explicit coordination~\cite{wen2022multi}, whereas recent advancements allow LLM-powered agents to communicate and collaborate through natural language, enabling human-like interactions~\cite{sclar2023minding,shanahan2023role}. Several systems, including VIRSCI~\cite{su2025headsbetteroneimproved} and SciAgents~\cite{ghafarollahi2024sciagents}, have adopted multi-agent frameworks to simulate research teams. These systems typically generate ideas through sequential discussions within fixed workflows. However, such designs fail to capture the stochastic, asynchronous, and multi-directional nature of real scientific discussions.
Specifically, VIRSCI underemphasizes intra-agent communication and does not model agent roles effectively, while SciAgents overlooks agent diversity, ignoring the varied expertise and perspectives that are essential to genuine collaboration. These limitations restrict the capacity of current systems to emulate the complexity of scientific practice.

To address these challenges, we propose IDVSCI (Internal Discussion and Vote SCIentists), a novel LLM-based multi-agent framework that improves idea generation through two key innovations: a Dynamic Knowledge Exchange mechanism and a Dual-Diversity Review paradigm. In the first stage, agents engage in cross-agent modification, contributing distinct perspectives to a proposed idea and overcoming the limitations of rigid sequential discussion. This is followed by collective aggregation, distilling the core insights into a refined idea. Finally, a dual-diversity review process assigns reviewers from varied knowledge backgrounds with tailored prompts to ensure multi-faceted evaluation. This design not only enhances the quality of assessment but also promotes inclusive and insightful review dynamics, potentially offering a blueprint for real-world scientific evaluation.
In addition to proposing a novel system, we introduce a new dataset in the health sciences domain, which enables us to assess the system's applicability beyond computer science and demonstrates its potential for cross-disciplinary scientific collaboration through multi-agent simulation.

To evaluate IDVSCI, we conduct extensive experiments on two datasets: a well-established benchmark in computer science, and our newly introduced health sciences dataset. IDVSCI achieves consistently superior performance across both, outperforming strong baselines such as AI Scientist and VIRSCI. Our results not only confirm the effectiveness of modeling agent interaction and diversity, but also uncover the potential of diverse collaboration mechanisms for advancing autonomous discovery.
Our main contributions are as follows:
\begin{enumerate}
\item 
We develop a dynamic ecosystem for idea generation in multi-agent systems by enabling cross-agent modification, collective aggregation, and iterative refinement. This framework empowers agents to contribute meaningfully and enhances overall idea quality, unlocking greater potential for innovation in complex tasks.
\item 
We propose a novel dual-diversity paradigm for the review process. By predefining scientists with diverse backgrounds and equipping them with context-specific prompts, our method improves the identification of truly exceptional ideas, offering insights into more effective scientific evaluation within multi-agent teams.
\item 
We further optimize the integration between dynamic knowledge exchange and the dual-diversity paradigm. This enhanced framework supports efficient configuration of multi-agent systems, leading to high-quality scientific ideation and laying the groundwork for autonomous discovery through collective intelligence.
\end{enumerate}

\section{Related Work}
\subsection{LLM-based Agents}
The advent of LLMs, exemplified by ChatGPT ~\cite{Chatgpt} and LLaMA ~\cite{dubey2024llama}, has revolutionized the capabilities of conversational agents, catalyzing a surge in research focused on LLM-based agents. These agents are sophisticated AI systems engineered to emulate specific individuals by leveraging the advanced functionalities of LLMs, such as command adherence, and social intelligence ~\cite{chen2024persona}. This enables them to effectively assist humans or autonomously execute tasks.

LLM-based single agents have been deployed in domains such as gaming, healthcare, and professional services ~\cite{agrawal2023multimodal,wozniak2024personalized}, functioning as game characters, medical consultants, or collaborative teammates~\cite{ijcai2024p711}.
Nevertheless, single-agent system often fall short when confronted with complex tasks ~\cite{chen2024auto}. To address this limitation, LLM-based multi-agent systems have been increasingly employed to tackle intricate challenges~\cite{chen2024agentverse}. These systems typically adopt two primary strategies~\cite{shang2024unified}: the first involves predefining multiple agent roles and allocating specific tasks to each, building upon the single-agent framework~\cite{qianetal2024chatdev,hong2024metagpt}. The second strategy employs dynamically defined roles, offering greater flexibility by allowing agents to determine tasks and roles based on factors such as memory and environmental cues~\cite{liuautonomous,gao2024360}. This dynamic approach enhances the adaptability and efficiency of multi-agent systems in complex scenarios.

\subsection{Agents in Scientific Research}
The pioneering work of AI-Scientist~\cite{lu2024ai} marked a significant milestone by exploring the potential of leveraging LLMs to simulate scientific research processes, positioning AI as a collaborative partner to human scientists. This approach enables AI agents to engage in open-ended scientific discovery, thereby assisting and, in some cases, replicating moments of human creativity and serendipitous innovation. Building on this foundation, VIRSCI~\cite{su2025headsbetteroneimproved} introduced a novel multi-agent framework enhanced by reinforcement learning, further advancing the generation of scientific ideas. By employing teams of agents, VIRSCI more accurately emulates real-world research environments compared to single-agent systems, fostering the generation of richer and more diverse ideas while promoting critical evaluation and innovation.
Novelseek~\cite{team2025novelseek} complements this direction by significantly improving the efficiency and precision of automated research through interactive human-AI collaboration.

Agent Laboratory~\cite{schmidgall2025agent} showcased the use of multi-agent systems to emulate collaborative scientific workflows, with agents capable of experimenting, analyzing, and evaluating ideas. This approach has since been applied across disciplines~\cite{liu2024harnessing,song2024multi}. For example, SciAgents~\cite{ghafarollahi2024sciagents} and AtomAgents~\cite{ghafarollahi2024atomagents} integrate materials science expertise for novel material discovery, while ProtAgent~\cite{ghafarollahi2024protagents} enables cross-domain protein design.
In contrast to prior approaches, our proposed system, IDVSCI, follows the same general workflow but introduces a novel strategy that simulates the behavioral dynamics of real-world scientific teams to generate more creative and diverse research ideas. As shown in Figure~\ref{fig_1}, our focus lies in facilitating early-stage ideation through agent-based collaboration, offering a distinct perspective on scientific discovery.

\section{Method}
\subsection{Task Definition}
Our task is to design a multi-agent system comprising \( n \) scientific agents, denoted as \( S_{\text{team}} = \langle \text{A}_1, \text{A}_2, \dots, \text{A}_n \rangle \). Through communication and collaboration within a predefined workflow, the team generates a scientific abstract, formally expressed as \( \mathit{Abstract} = f(S_{\text{team}}) \), where \( f(\cdot) \) denotes the  collaborative generation process.

\subsection{Dynamic Knowledge Exchange}

\begin{figure*}[!t]
    \centering
    \includegraphics[width=0.9\textwidth]{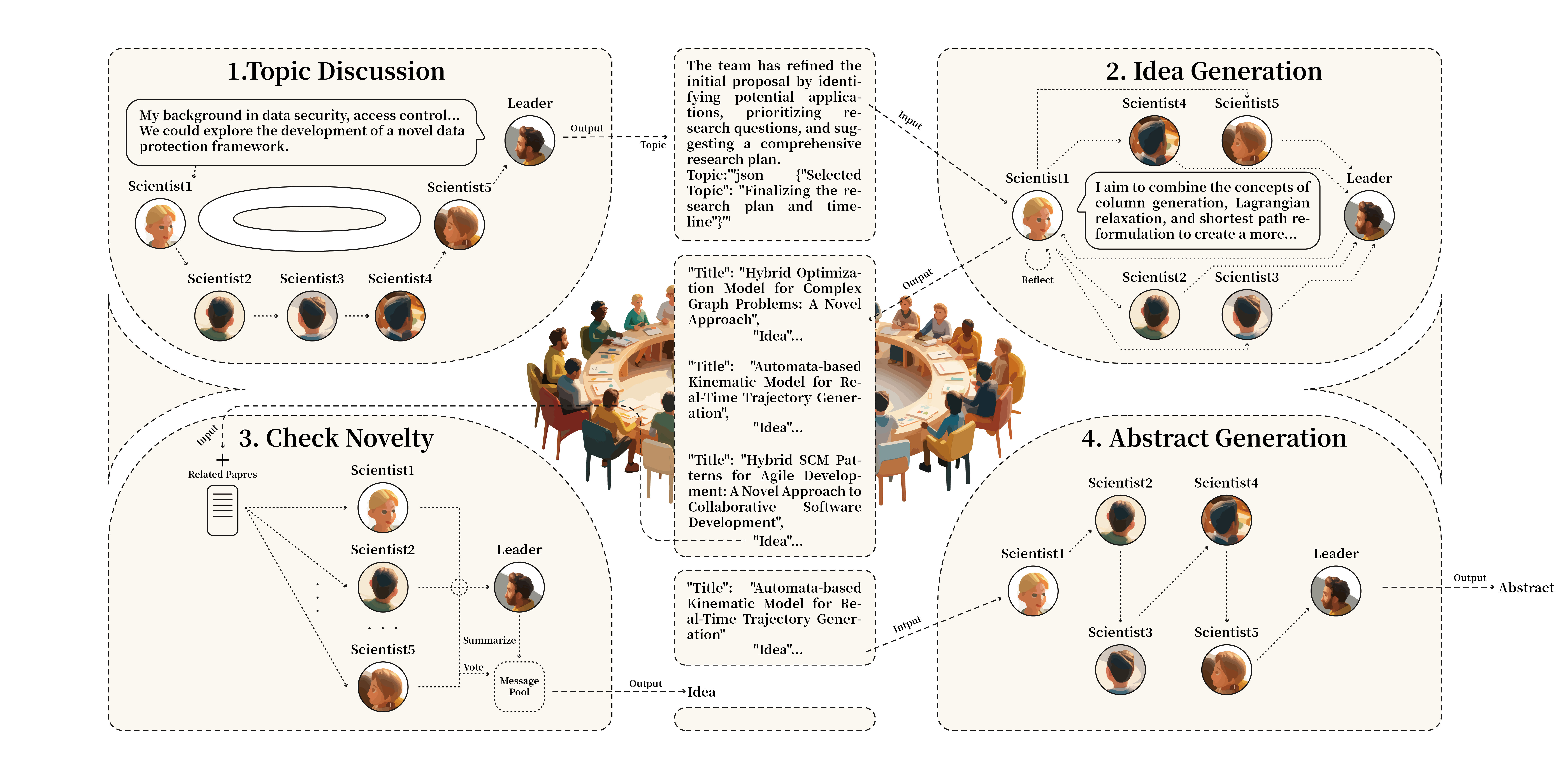}
    \caption{The workflow framework of our method designed for automated scientific research. It consists of four sequential steps: Topic Discussion, Idea Generation, Check Novelty, and Abstract Generation. The line segments represent the transmission of information, with the direction of the arrows indicating the recipient agents.}
    \label{fig_2}
    \end{figure*}

In conventional multi-agent systems, LLM-based agents are typically assigned static roles and communicate sequentially within a fixed workflow, generating responses based on dialogue history. While effective under shared knowledge conditions, this setup becomes problematic when agents possess distinct knowledge bases, as is often the case in automated scientific research. In such scenarios, the rigid communication structure leads to inefficient knowledge sharing and repetitive outputs, particularly when model capacity is limited. Additionally, unstructured turn-taking often results in long dialogue histories, further degrading response quality.

To address these limitations, we propose a Dynamic Knowledge Exchange (DKE) mechanism, inspired by the review workflow used in major AI conferences, specifically the Area Chair (AC) and Program Committee (PC) model. Unlike the traditional sequential communication approach, our method introduces a leader role, analogous to the AC in conference reviews, while the other agents assume roles similar to PC. Specifically, each ordinary scientist, equipped with its unique knowledge base and prompt, is responsible for generating and presenting its ideas to the leader. The leader, leveraging its own knowledge and functional expertise, aggregates and synthesizes the outputs from all team members into a cohesive summary.

Formally, let \( \mathcal{A} = \{A_1, A_2, \dots, A_n\} \) denote the set of agents, where \( A_1 \) represents the leader and \( A_2, \dots, A_n \) represent the scientists, such that each scientist \( A_i \) (\( i \geq 2 \)) generates an idea \( I_i = \text{generate}(K_i, P_1) \) based on their respective knowledge base \( K_i \) and a given prompt \( P_1 \).

These ideas are then cross-reviewed by the other scientists. That is, for each idea \( I_i \), another agent \( A_j \) (\( j \neq i \)) produces a revised version: \( I'_{i,j} = \text{revise}(K_j, P_2, I_i) \), where \( P_2 \) is a secondary prompt used during peer review.

The leader \( A_1 \) then synthesizes all revised ideas for each initial idea \( I_i \) into a coherent summary:

\begin{equation}
S_i = g\left( \{I'_{i,2}, I'_{i,3}, \dots, I'_{i,i-1}, I'_{i,i+1}, \dots, I'_{i,n}\}, K_1 \right)
\end{equation}
where \( g(\cdot) \) denotes the synthesis function, and \( K_1 \) denotes the leader's knowledge base. 

Compared to standard sequential communication, our DKE method offers several advantages. First, it reduces redundancy by avoiding excessive reliance on long dialogue histories. Second, it enriches idea diversity through structured cross-agent review. Third, it better reflects real-world scientific workflows, where a central coordinator synthesizes insights from diverse contributors. By promoting structured, role-aware interaction and leveraging heterogeneous knowledge, DKE enhances both the efficiency and creativity of scientific idea generation in multi-agent LLM systems.

\subsection{Dual-Diversity Review}

To simulate a realistic scientific research environment, we introduce the Dual-Diversity Review (DDR) mechanism, which incorporates agents with diverse knowledge backgrounds and dynamically updated prompts. This design not only captures the heterogeneity of real-world research teams, but also promotes innovation by integrating varied perspectives and domain knowledge.

Prior to initiating automated research, the environment is configured with two core components: a set of scientists and a collection of prior publications. The latter, termed the \(Past Paper Dataset\), serves as an external knowledge source throughout the research process. Let $\mathcal{M}$ denote the adjacency matrix representing prior collaborations among scientists, where \(M_{i,j} \) indicates the number of co-authored papers between \(A_i \) and \(A_j \).  Following the strategy of ~\cite{su2025headsbetteroneimproved}, we mitigate the bias toward selecting familiar collaborators—an issue shown to suppress novelty~\cite{zeng2021fresh}—by adding 1 to all entries in $\mathcal{M}$ to encourage the formation of novel team compositions.
To further maximize diversity, we construct teams with partially overlapping knowledge domains, allowing agents to introduce complementary ideas during discussion. The overall team diversity is quantified as:
\vspace*{-0.5em}
\begin{equation}
\text{Diversity}(S_{team}) = \sum_{i=1}^n \sum_{j=i+1}^n \text{Distance}(K_i, K_j)
\end{equation}

where \( \text{Distance}(K_i, K_j) \) measures the dissimilarity between the knowledge bases of agents \( A_i \) and \( A_j \).

During the idea generation phase, the DDR mechanism continuously refines each agent's prompts by integrating the most relevant literature. Specifically, we leverage the Faiss library~\cite{douze2024faiss} to compute the Euclidean distances between each generated idea and papers in the database.

The top-\( k \) nearest papers are retrieved as reference material, allowing agents to ground their ideas in current scientific contexts. Formally, the updated prompt \( P_i' \) for scientist \( A_i \):
\vspace*{-0.3em}
\begin{equation}
P_i' = P_i \cup \{Paper_j \mid Paper_j \in \text{Top-}\textit{k}(D(I, Paper))\}
\end{equation}

where \( \text{Top-}\textit{k}(D(I, Paper)) \) returns the \( k \) papers closest to idea \( I \) based on Euclidean distance.

This approach ensures that the diversity of prompts through dynamically updated and continuously optimized to enhance creativity and impact, rather than remaining static or requiring manual intervention. By establishing a set of general rules, the system autonomously refines the prompts through dialogues with other scientists and the exploration of relevant literature. This process closely mirrors the way real-world researchers acquire new ideas, making the DDR mechanism both efficient and highly realistic.

\subsection{Framework}
Similar to the work of~\cite{su2025headsbetteroneimproved}, after configuring the scientist team, we divide the workflow for abstract generation into four distinct stages: Topic Discussion, Idea Generation, Check Novelty, and Abstract Generation. Among these, Idea Generation and Check Novelty are the most critical phases that determine the innovativeness of the final ideas.

\textbf{Topic Discussion} In this initial stage, we follow the approach of ~\cite{su2025headsbetteroneimproved}, where each agent in the team generates potential topics based on a predefined prompt, its own background knowledge, and the sequential dialogue history of the team.
The team then selects a final topic from the proposed topics based on a probability distribution, ensuring a balanced and representative choice.This stage ensures that the team converges on a well-defined and relevant topic, laying the foundation for the subsequent phases of idea generation and novelty checking.

\textbf{Idea Generation}
Existing methods in the idea generation phase fail to adequately simulate real-world scientific discussions. These approaches typically employ a process similar to the previous Topic Discussion stage, where agents take turns generating ideas based on predefined prompts. Each agent combines its background knowledge and dialogue history to refine or generate new ideas, and the team ultimately selects the most innovative ideas based on a scoring mechanism. However, this approach lacks the collaborative and interactive nature of real-world scientific teams, particularly the brainstorming sessions that are crucial for fostering creativity.

To fully leverage the diversity of agent backgrounds, we introduce the DKE method. Our design is as follows: After generating an initial idea \( I_i \), each scientist \( \text{A}_i \) shares \( I_i \) with all other scientists except the leader. Each receiving scientist then modifies \( I_i \) based on their unique background and knowledge. These modified ideas are submitted to the leader, who synthesizes the feedback along with their own knowledge to produce a consolidated revision. This revision is returned to the original scientist \( \text{Scientist}_i \), who reflects on the feedback and refines \( I_i \) into a final version \( I'_i \).

Formally, let \( I_i \) denote the initial idea generated by Scientist \( \text{A}_i \), and let \( I'_{i,j} \) represent the modification made by scientist \( \text{A}_j \) to \( I_i \). The leader aggregates these modifications into a unified revision \( S_i \), which is then used by scientist \( \text{A}_i \) to produce the final idea \( I'_i \):
\vspace*{-0.3em}
\begin{equation}
I'_i = \text{Reflect}(I_i, A_i)
\end{equation}

where \( \text{Reflect}(\cdot) \) represents the reflection process that integrates the leader's feedback into the initial idea.

\begin{figure}[!t]
    \centering
    \includegraphics[width=\columnwidth]{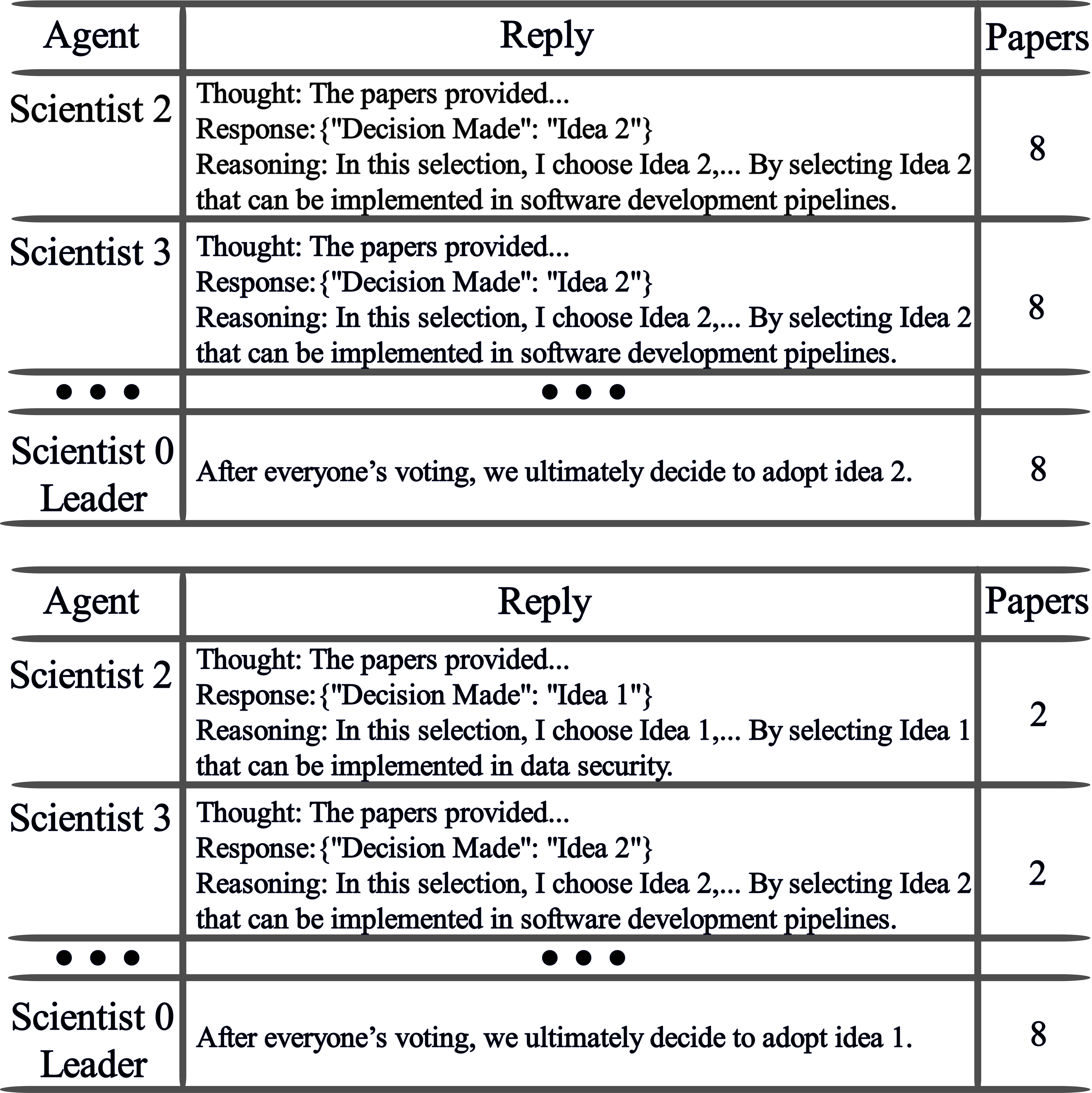}
    \caption{A case study illustrates the responses generated by agents utilizing the same LLM when provided with 8 and 2 references, respectively, as well as the final output results produced by the multi-agent system.}
    \label{fig_3}
\end{figure}

\textbf{Check Novelty}
Through empirical experimentation, we observed that during the novelty assessment phase, different scientists occasionally generate identical responses for the same idea, producing highly similar \textit{thoughts} and \textit{reasonings}. Upon removing a subset of the reference papers, we noted a significant reduction in this phenomenon. As illustrated in Figure~\ref{fig_3}, an increase in the number of reference papers tends to result in agents with different backgrounds generating identical outputs. Conversely, reducing the number of reference documents leads to more diverse and distinct outputs. This observed relationship highlights the critical importance of optimizing both the quantity and quality of reference materials to enhance the variability and originality of the generated content.

To address the challenges posed by the large volume of reference papers and to ensure comprehensive coverage without omission, we evenly distributed the related papers among all scientists. This approach ensures that each scientist evaluates a manageable subset of the literature, while collectively covering the entire body of relevant work. To further integrate the diverse perspectives of these scientists with varied backgrounds in a fair and systematic manner, we employ a weighted Borda count voting mechanism. This method is particularly suitable for ranking ideas based on their novelty, feasibility, and potential impact, as it allows for the aggregation of rankings and confidence scores from multiple evaluators. The Borda count was chosen because it not only accounts for the relative rankings of ideas but also incorporates the evaluators' confidence in their assessments, ensuring a balanced and nuanced evaluation process.

Let \( I = \{i_1, i_2, \dots, i_n\} \) represent the set of ideas, where each idea \( i_k \) is assigned a score based on the voting procedure. Let \( V_j \) denote the votes from the \( j \)-th scientist, where \( V_j \) is a tuple consisting of the rankings \( r_{jk} \) of the ideas and the confidence scores \( c_{jk} \) associated with these rankings. The Borda score \( B_k \) for idea \( i_k \) is computed as follows:
\vspace*{-0.3em}
\begin{equation}
B_k = \sum_{j=1}^{m} \left( (n - r_{jk}) \times \frac{c_{jk}}{10} \right)
\end{equation}

where \( n \) is the total number of ideas, \( r_{jk} \) is the rank assigned to idea \( i_k \) by scientist \( j \), \( c_{jk} \) is the confidence score given by scientist \( j \) for idea \( i_k \), and \( m \) is the total number of scientists.
The idea with the highest Borda score \( B_k \) is selected as the final choice, reflecting the team's consensus on the novelty and feasibility of the idea.

\textbf{Abstract Generation}
In this final phase, since the ideas have already been generated, our primary objective is to ensure that the quality of the generated abstract is solely influenced by the innovativeness of the ideas. To achieve this, we adopt a methodology consistent with previous approaches. Specifically, each scientist generates an initial draft of the abstract based on the finalized ideas. Subsequently, the scientists take turns sequentially refining the abstract. This iterative modification process ensures that the final output remains closely tied to the quality of the underlying ideas, minimizing external influences on the abstract's content.
By employing this method, we ensure that the quality of the generated abstract is primarily determined by the innovativeness and clarity of the ideas themselves, rather than being influenced by extraneous factors. This approach not only maintains the integrity of the abstract generation process but also aligns with the goal of producing high-quality, idea-driven summaries that accurately reflect the scientific contributions of the team.

\section{Experiment}
\begin{table*}[htbp]
\centering
\caption{Comparison of our model with state-of-the-art baselines experimental results}
\label{table_1}
\begin{tabular}{ccccc|cccc}
\hline
\multicolumn{1}{l}{}    & \multicolumn{4}{c|}{Computer Sciences}                        & \multicolumn{4}{c}{Health Sciences}                            \\ \hline
Model                   & HD↑           & CD↓           & CI↑           & ON↑           & HD↑           & CD↓           & CI↑           & ON↑            \\ \hline
AI-Scientist: LLAMA-8b  & 0.51          & 0.49          & 2.12          & 2.21          & -             & -             & -             & -              \\
AI-Scientist: LLAMA-70b & \textbf{0.53} & 0.48          & 2.11          & 2.33          & -             & -             & -             & -              \\
VIRSCI: LLAMA-8b        & 0.43          & 0.42          & 3.29          & 3.40          & 0.39          & \textbf{0.37} & 7.70          & 8.12           \\
VIRSCI: QWQ-32b         & 0.46          & 0.40          & 3.38          & 3.89          & 0.40          & 0.38          & 7.82          & 8.23           \\
VIRSCI: LLAMA-70b       & 0.44          & 0.40          & 3.36          & 3.70          & 0.39          & \textbf{0.37} & 7.84          & 8.26           \\
IDVSCI: LLAMA-8b        & 0.40          & \textbf{0.39} & 4.38          & 4.49          & 0.40          & 0.39          & 8.37          & 8.58           \\
IDVSCI: QWQ-32b         & 0.41          & 0.40          & 4.17          & 4.27          & \textbf{0.42} & 0.40          & 8.09          & 8.49           \\  
IDVSCI: LLAMA-70b       & 0.40          & \textbf{0.39} & \textbf{4.49} & \textbf{4.60} & 0.41          & 0.39          & \textbf{10.06}& \textbf{10.58} \\ \hline
\end{tabular}
\end{table*}
\subsection{Dataset}
For our experimental framework, we employed the publicly available Computer Sciences Dataset, derived from the AMiner\footnote{\url{https://www.aminer.cn/aminernetwork}} computer science repository~\cite{tang2008arnetminer}. The final curated dataset comprised 156 researchers and 85,217 publications, which served as the foundation for ecosystem construction and agent initialization. All publication and researcher metadata were processed using the mxbai-embed-large embedding model~\cite{lee2024open} for feature representation.
Inspired by Novelseek~\cite{team2025novelseek}, which demonstrates the effectiveness of human-AI collaboration in multidisciplinary scientific applications, we further introduced a new dataset—the Health Sciences Dataset—to evaluate the cross-domain scalability of our method in promoting innovative idea discovery. This dataset includes 130 researchers and 86,448 publications, collected from PubMed, and follows the same preprocessing pipeline as the Computer Sciences Dataset for ecosystem construction and agent initialization.

\subsection{Experimental Setup}
The proposed framework is developed utilizing the Agentscope architecture~\cite{gao2024agentscope}, a robust platform designed for constructing multi-agent systems powered by LLMs. For performance evaluation, we employ open-source LLMs with varying computational scales, including LLaMA-3.1 models with 8b and 70b parameters~\cite{dubey2024llama}, as well as the recently released QWQ-32b model~\cite{qwq32b}. All models are accessed through the Ollama interface and executed on four NVIDIA RTX 4090 GPUs. This diversity of backbone models enables us to assess the performance of our framework across different levels of language model capacity.
Mirroring the experimental protocol established in VIRSCI, our implementation features a four-agent system conducting iterative discussions across five sequential rounds. To ensure statistical reliability, all performance metrics are derived from an average of 20 independent experimental trials.
To balance the number of reference papers and avoid an excessive or insufficient selection, we empirically set \( k = 8 \) for \(\text{Top-}\textit{k}(D(I, \text{Paper}_j))\) in the Check Novelty phase.
We conducted experiments on both datasets when evaluating the state-of-the-art baselines. All subsequent experiments were conducted solely on the Computer Sciences Dataset.

\subsection{Evaluation Metrics}
Currently, no single evaluation metric can fully capture the innovativeness of scientific outputs. In line with VIRSCI's methodology, we utilize four established metrics to provide a partial but insightful evaluation of scientific innovation.

1. Historical Dissimilarity (HD): Defined as the average squared Euclidean distance between the embedding vector of the generated abstract and those of the 5 most similar abstracts in the corpus of pre-2011 literature \cite{shao2020bert,zhou2024fine}. A larger distance indicates higher dissimilarity from historical works, reflecting greater potential for innovation.

2. Contemporary Dissimilarity (CD): Calculated as the average squared Euclidean distance between the embedding vector of the generated abstract and those of the 5 most similar abstracts in the corpus of post-2011 literature. A smaller distance suggests higher similarity to recent works, also indicating greater potential for innovation.

3. Contemporary Impact (CI): This reflects the citation count of the top 5 most similar abstracts in the corpus of articles published after 2011~\cite{yang2022gender}. A higher citation count suggests that the generated abstract is likely to have a greater impact.

To ensure comparability, each calculated metric is normalized using the mean value derived from the entire corresponding database, with normalization defined as the metric divided by its mean value.

4. Overall Novelty (ON): ON is positively correlated with HD and CI and negatively correlated with CD. It is calculated as \( \text{ON} = \frac{\text{HD} \times \text{CI}}{\text{CD}} \). Mathematically, the expected value of ON is proportional to the true novelty.


\subsection{Results}

To ensure fair comparison, we evaluate all methods using models with aligned parameter scales and similar architectures. As shown in Table~\ref{table_1}, our method IDVSCI consistently outperforms AI-Scientist and VIRSCI across all metrics in both Computer Sciences and Health Sciences domains.
In the Computer Sciences domain, IDVSCI (LLaMA-70b) achieves the highest CI and ON, while maintaining competitive HD and CD, indicating its ability to generate both novel and scientifically valuable ideas. In Health Sciences, IDVSCI again shows strong performance, achieving the highest CI (10.06) and ON (10.58), demonstrating its cross-domain adaptability and capacity to produce impactful, domain-relevant content.

While larger models generally yield better CI and ON, we observe that IDVSCI with QWQ does not always outperform its LLaMA-8b variant, suggesting that increasing model size alone does not guarantee performance gains in more complex multi-agent settings.
Overall, these results demonstrate the effectiveness and robustness of IDVSCI, while offering insights into the interplay between model scale, output complexity, and system design in scientific idea generation.
\subsection{Ablation Experiments}
\textbf{Module-wise Ablation Study}
To evaluate the contribution of each component, we conducted ablation experiments (Table~\ref{table_2}). Removing the internal discussion module in the Idea Generation stage leads to the most significant performance degradation, with CI dropping from 4.38 to 4.10 (-6.4\%). This confirms that structured discussions substantially enhance the impact of generated ideas. 
Interestingly, removing the voting mechanism in the Check Novelty stage slightly increases CI (4.38 → 4.42, +0.9\%), but at the expense of ON, which decreases from 4.49 to 4.42. This suggests that while voting may sometimes constrain idea impact, it plays an essential role in preserving originality by avoiding over-convergence in novelty evaluation. 
\begin{table}[htbp]
    \centering
    \caption{Comparison results between our method and the method of removing different modules in it. Where ``-'' denotes the removal of the module.}
    \label{table_2}
    \begin{tabular}{cccccc}
    \hline
                                                & \multicolumn{1}{c}{HD↑}  & \multicolumn{1}{c}{CD↓}  & \multicolumn{1}{c}{CI↑}  & \multicolumn{1}{c}{ON↑}  \\ \hline
    - Team Discuss              & \textbf{0.41}               & 0.38                     & 4.10                     & 4.42                     \\
    \multicolumn{1}{l}{- Team Vote} & 0.38         &  \textbf{0.38}           & 4.42                     &  4.42                    \\
    IDVSCI                                      & \multicolumn{1}{c}{0.40} & \multicolumn{1}{c}{0.39} & \textbf{4.38} & \textbf{4.49} \\ \hline
    \end{tabular}
    \end{table}

\textbf{Iteration-wise Ablation Study}
The number of discussion turns is a critical factor influencing inference cost, as each additional round requires further computation and communication among agents. As shown in Table~\ref{table_3}, the creativity and impact of ideas do not vary substantially with different iteration counts. Specifically, ON values exhibit very low variance (0.014), with a mean of 4.46 and a maximum deviation of only 0.16. Similarly, CI scores remain stable, ranging narrowly between 4.18 and 4.45. These results suggest that a single iteration already provides strong performance (ON = 4.61, CI = 4.38), while additional turns bring limited or inconsistent gains (e.g., CI drops to 4.18 at iteration 4). From a cost-effectiveness perspective, our framework achieves near-optimal outcomes with just 1--2 iterations, underscoring its efficiency in balancing creativity, novelty, and computational overhead.

\begin{table}[]
\centering
\caption{Comparison of results generated by our method across different rounds of discussion.}
\label{table_3}
\begin{tabular}{cllll}
\hline
Iteration & HD↑  & CD↓  & CI↑  & ON↑  \\ \hline
1         & 0.40 & 0.38 & 4.38 & 4.61 \\
2         & 0.39 & 0.38 & 4.24 & 4.35 \\
3         & 0.41 & 0.40 & 4.45 & 4.56 \\
4         & 0.38 & 0.37 & 4.18 & 4.30 \\
5         & 0.40 & 0.39 & 4.38 & 4.49 \\ \hline
\end{tabular}
\end{table}

\subsection{Impact of Agent Background Diversity}
\begin{figure}[!t]
\centering
\includegraphics[width=\columnwidth]{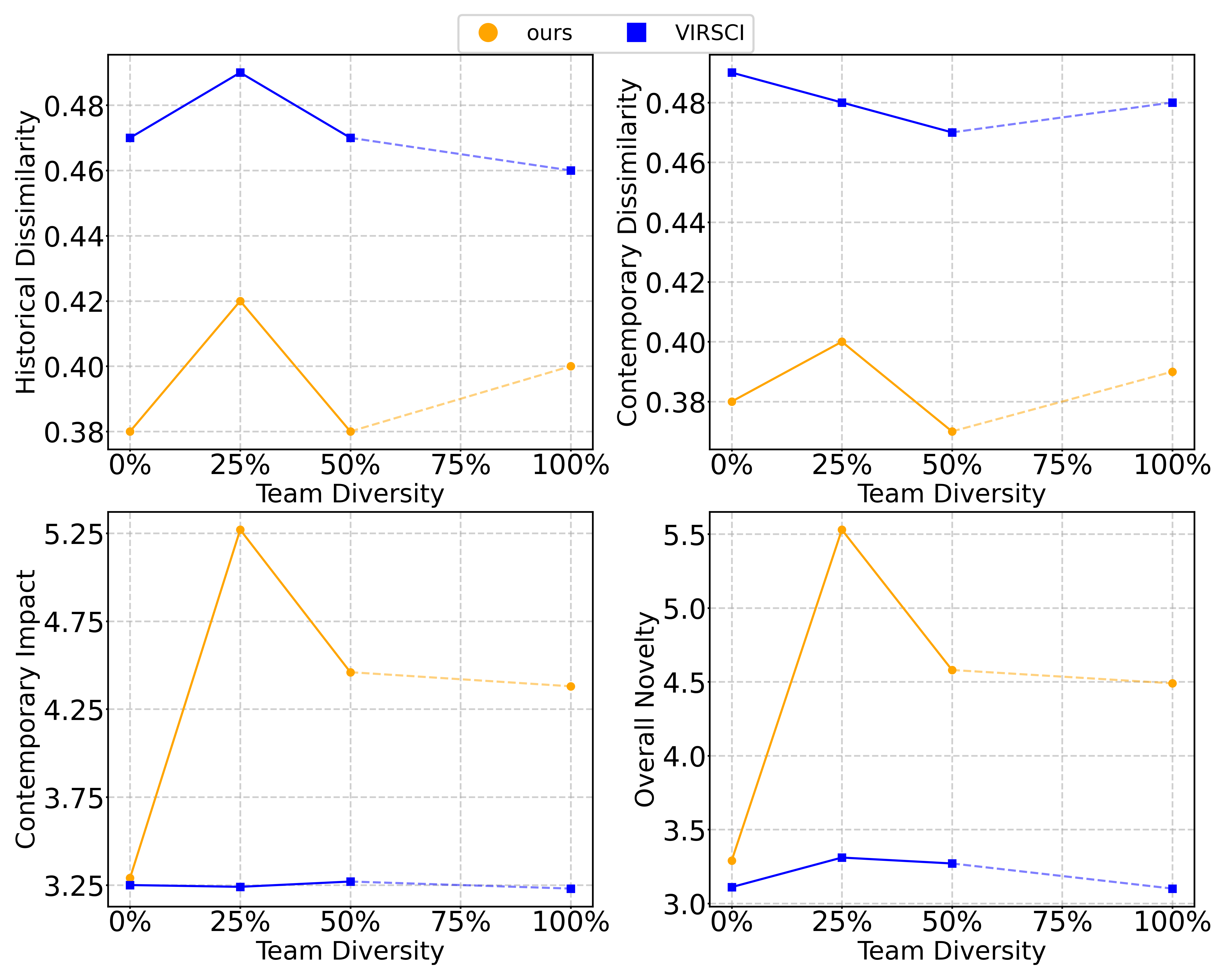}
\caption{Comparative experimental data for visualizing generated results under different levels of background diversity.}
\label{fig_4}
\end{figure}

As shown in Figure~\ref{fig_4}, when the backgrounds of the scientists constituting the research team are entirely homogeneous, the generated ideas perform the poorest in terms of both innovativeness and impact. The model achieves optimal performance when the background diversity reaches 25\%. At 50\% and 100\% diversity levels, the effects are comparable, though slightly inferior to the former. This aligns closely with our understanding of real-world research teams.

\subsection{Impact Analysis of Our Method on Internal and External Teams}
\begin{table}[tbp]
    \centering
    \caption{Comparison of external vs. internal teams using our methodology.}
    \label{table_4}
    \begin{tabular}{cllll}
    \hline
                               & HD↑ & CD↓ & CI↑  & ON↑  \\ \hline
    ERSCI                      & 0.40 & 0.39 & 4.04 & 4.14 \\
    \multicolumn{1}{l}{ours} & 0.40 & 0.39 & \textbf{4.38} & \textbf{4.49} \\ \hline
    \end{tabular}
    \end{table}
    As illustrated in Table~\ref{table_4}, the application of our method to generate summaries with the assistance of external teams-External Review SCIentists (ERSCI)-results in inferior outcomes compared to those produced through internal collaborative discussions. While both approaches achieve identical HD and CD values, IDVSCI demonstrates a superior performance in CI. This discrepancy can be primarily attributed to the limited communication between the external method and the originators of ideas, which leads to information distortion during the transmission process.

\subsection{Impact of Team Size}
\begin{figure}[!t]
    \centering
    \includegraphics[width=0.5\columnwidth]{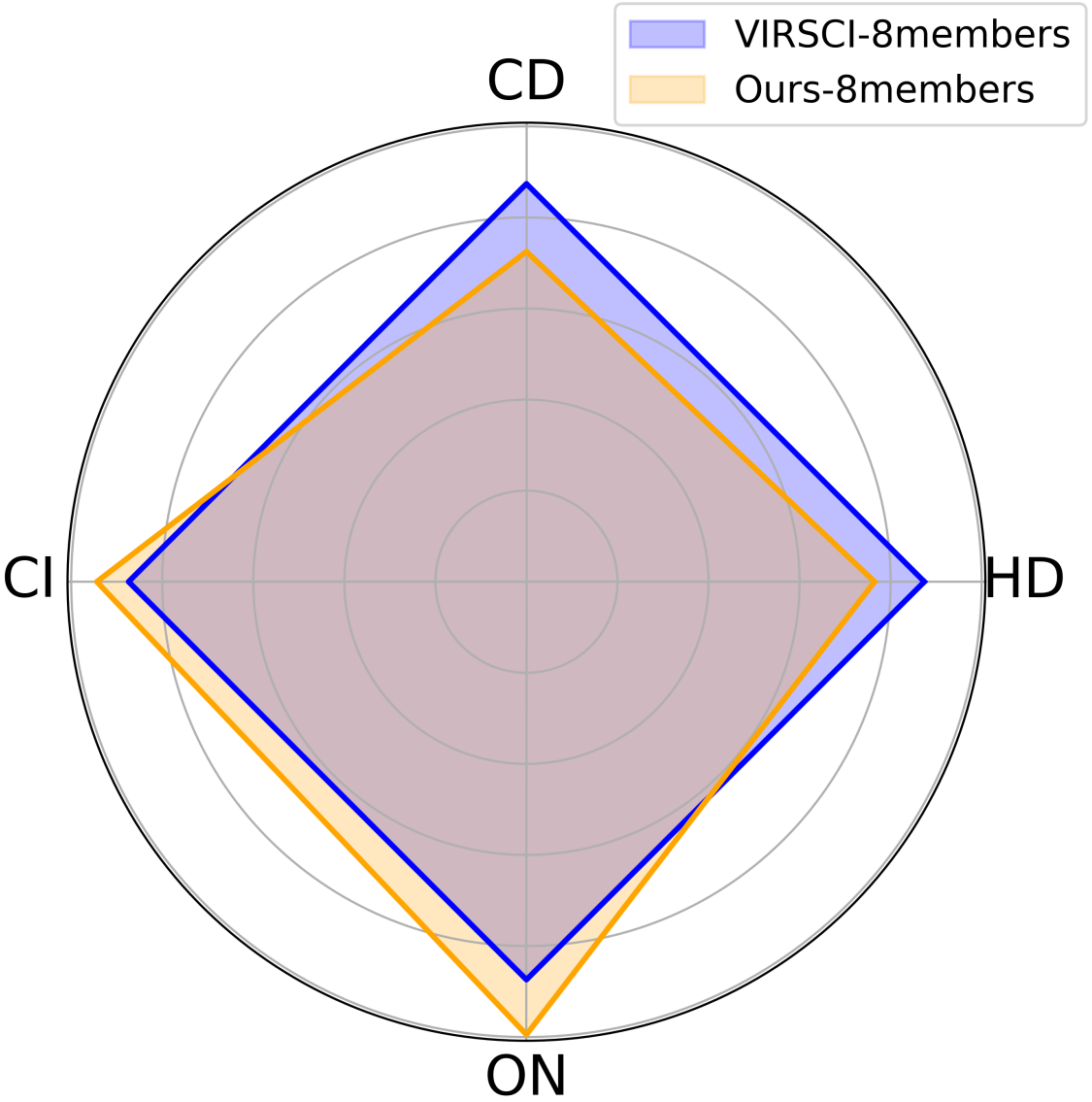}
    \caption{Comparison chart of results when team size is 8.}
    \label{fig_5}
    \end{figure}
   
The experimental results from the VIRSCI team indicate that their multi-agent research team performs optimally when the team size is 4 and 8. In previous comparisons with the SOTA models, to avoid unfair comparisons with methods like AI-Scientist due to excessively large team sizes, we only compared the results for a team size of 4. As can be seen from Figure~\ref{fig_5}, our method still achieves the best performance at a team size of 8. However, compared to VIRSCI, our method does not show significant improvement when scaling from 4 to 8 members, suggesting that our approach can achieve desirable outcomes without incurring substantial additional costs.

\section{Conclusion}
By establishing a dynamic knowledge-exchange ecosystem that incorporates cross-agent modification, collective aggregation, and iterative refinement, alongside a novel dual-diversity paradigm for the review process, we enhance both the idea-generation and evaluation mechanisms within multi-agent systems. Furthermore, we introduce a new dataset in the health sciences domain, enabling the exploration of multi-agent scientific systems in cross-disciplinary settings. This research paves the way for more faithful emulation of real-world scientific practices, more inclusive and rigorous peer review, and ultimately, deeper scientific breakthroughs—by unlocking the collective intelligence of multi-agent systems and advancing diverse modes of collaborative research.


\bibliography{aaai2025}

\appendix
\onecolumn
\begin{tcolorbox}[title={\faBook\quad Health Sciences Dataset\quad\faStethoscope}, myfancytitle, breakable]
    \numberedbox[\faBrain\ Health Sciences Author]{
        Your name is Scientist3, you belong to following affiliations [\textquotesingle Department of Dermatology, Health New Zealand Te Whatu Ora Auckland, New Zealand.\textquotesingle], you have researched on following topics [\textquotesingle*Vulvar Diseases/drug therapy\textquotesingle, \textquotesingle*Hydroxychloroquine/therapeutic use/adverse effects/administration \& dosage\textquotesingle, \textquotesingle Female\textquotesingle, \textquotesingle Treatment Outcome\textquotesingle, \textquotesingle*Lichen Planus/drug therapy\textquotesingle], you have published 1 papers,you have previously collaborated with these individuals [\textquotesingle Scientist4\textquotesingle, \textquotesingle Scientist5\textquotesingle].
    }    
    \numberedbox[\faFile\ Health Sciences Past Paper Dataset]{
    \textbf{Title:} The empirical testability of Skog's theory of collective drinking behaviour.\\
    \textbf{PMID:} 28474448\\
    \textbf{Authors:} Gmel G, Rehm J\\
    \textbf{Affiliations:} Swiss Institute for the Prevention of Alcohol and other Drug Problems, Lausanne, Switzerland; Social Prevention and Health Policy Research Department, Addiction Research Foundation, Toronto, Canada; Department of Public Health Sciences, Faculty of Medicine, University of Toronto, Canada; WHO, Geneva, Switzerland.\\
    \textbf{Year:} 2000\\
    \textbf{Venue:} Drug and alcohol review\\
    \textbf{Citations:} 7\\
    \textbf{Abstract:} The objects of this study were (1) to review systematically Skog's theory of collective drinking behaviour and its interpretations by alcohol researchers, and (2) to give examples of how Skog's theory and these different interpretations have been empirically tested and to indicate how they might be tested. Based on a computer-aided search of the literature, a reconstruction of the theory and possible alternative interpretations is provided. Different interpretations of Skog's theory are possible and can be found in the literature. Surprisingly, there is little empirical evidence, especially recent evidence, to support Skog's key assumptions. Suggestions for further research are given.
    }
    \numberedbox[\faMagic\ Health Sciences Future Paper Dataset]{
    \textbf{Title:} Chronic Physical Health Problems in Sexual Minority Women: Review of the Literature.\\
    \textbf{PMID:} 26789854\\
    \textbf{Authors:} Eliason MJ\\
    \textbf{Affiliations:} Department of Health Education, San Francisco State University , San Francisco, California.\\
    \textbf{Year:} 2014\\
    \textbf{Venue:} LGBT health\\
    \textbf{Citations:} 12\\
    \textbf{Abstract:} Although there is substantial literature about sexual minority women's mental health and use of alcohol, tobacco, and other drugs (ATOD), only recently has attention been focused on chronic physical health disorders thought to stem from stress and exposure to ATOD use. The most extensively studied aspect of physical health has been weight, with the majority of studies reporting higher prevalence of overweight and obesity. In addition, many studies report higher levels of stressful experiences in both childhood and adulthood for sexual minority women. In this paper, the hypothesized relationship between stress, unhealthy behaviors, and five common chronic physical health disorders is explored via review of the literature. Only asthma appeared to be consistently more common in sexual minority women, and few or no differences in diabetes, hypertension, cardiovascular disease, and most cancers were found. The limitations of these studies are reviewed, and the need for studies that directly address the relationships among stress, health-damaging practices, and chronic disorder is emphasized. However, if these findings hold up, and sexual minority women are not more prone to these disorders, the field may need better theoretical frameworks from which to explore potential differences in the manifestation of mental versus chronic physical health disparities.
    }
\end{tcolorbox}
Figure 6: Health Sciences Dataset as the foundation for ecosystem construction and agent initialization.

\begin{tcolorbox}[title={Generated Abstract(IDVSCI: LLAMA-70b(ours))}, myfancytitle, breakable]    
    \numberedbox{
    \textbf{Title:} Fostering Resilience in Breast Cancer Patients Undergoing Reconstructive Surgery: A Novel Gratitude-Mindfulness Intervention with Neuro-Emotional Technique (NET) and Loving-Kindness Meditation (LKM)\\
    \textbf{Abstract:} Breast cancer patients undergoing reconstructive surgery often experience significant psychological distress, compromising their resilience and overall well-being. Despite advances in medical care, there remains a critical need for innovative interventions that promote resilience and self-care in this vulnerable population. Recent studies have highlighted the importance of gratitude and mindfulness practices in enhancing neural plasticity, opportunity recognition, and physical health outcomes.

    This randomized controlled trial aims to evaluate the efficacy of an 8-week novel gratitude-mindfulness intervention combined with Neuro-Emotional Technique (NET) sessions and Loving-Kindness Meditation (LKM) in enhancing neural plasticity, opportunity recognition, physical health outcomes, and psychological well-being in breast cancer patients undergoing reconstructive surgery. The study also explores the potential benefits of integrating Eastern Body-Mind-Spirit (BMS) Group Intervention model, mindfulness-based stress reduction (MBSR), LKM, and NET to support the complex needs of this population.

    A total of 120 breast cancer patients will be recruited and randomly assigned to either an intervention group or a control group. The intervention group will participate in weekly sessions combining gratitude practices, mindfulness exercises, LKM, and NET sessions to address cancer-related traumatic stress symptoms. Outcome measures include functional magnetic resonance imaging (fMRI) scans, surveys assessing opportunity recognition, standardized questionnaires evaluating physical health outcomes, psychological well-being, quality of life, and social support networks.

    Preliminary findings indicate significant improvements in neural plasticity, opportunity recognition, and physical health outcomes among participants in the intervention group compared to the control group. Notably, fMRI scans reveal increased activity in brain regions associated with emotional regulation and resilience. Participants in the intervention group also report enhanced psychological well-being, reduced stress levels, and improved quality of life. Furthermore, results suggest that the integration of NET and LKM enhances the effectiveness of the gratitude-mindfulness intervention.

    This study provides preliminary evidence for the efficacy of a novel gratitude-mindfulness intervention combined with NET and LKM in promoting resilience and enhancing physical health outcomes in breast cancer patients undergoing reconstructive surgery. The findings have important implications for the development of innovative interventions that integrate Eastern Body-Mind-Spirit (BMS) Group Intervention model, mindfulness-based stress reduction (MBSR), LKM, and NET to support the complex needs of this population. Future studies should aim to replicate these results and explore the long-term benefits of this intervention.

    Our study contributes to the existing literature by introducing a novel gratitude-mindfulness intervention that combines multiple evidence-based approaches to promote resilience in breast cancer patients undergoing reconstructive surgery. The integration of NET and LKM with gratitude practices and mindfulness exercises provides a comprehensive framework for addressing the complex needs of this population. By exploring the neural mechanisms underlying the intervention's effects, we can gain a deeper understanding of how these approaches influence emotional regulation, opportunity recognition, and physical health outcomes.

    The results of this study have significant implications for clinical practice, highlighting the potential benefits of incorporating gratitude-mindfulness interventions into standard care for breast cancer patients undergoing reconstructive surgery. By promoting resilience and enhancing physical health outcomes, these interventions can improve quality of life and reduce healthcare costs associated with chronic stress and anxiety.

    Moreover, our research highlights the importance of considering the interplay between psychological, social, and spiritual factors in the development of effective interventions for this population. The integration of NET and LKM with gratitude practices and mindfulness exercises provides a unique opportunity to explore the complex relationships between these factors and their impact on resilience and physical health outcomes.

    In conclusion, our study demonstrates the potential benefits of a novel gratitude-mindfulness intervention combined with NET and LKM in promoting resilience and enhancing physical health outcomes in breast cancer patients undergoing reconstructive surgery. The findings of this study have important implications for clinical practice and highlight the need for further research into the development of innovative interventions that address the complex needs of this population.\\
    \textbf{HD:} 0.31; \textbf{CD:} 0.34; \textbf{CI:} 17.63
}
\end{tcolorbox}
Figure 7: Example abstract generated by our IDVSCI model, focusing on psychological rehabilitation and intervention for breast cancer patients. 

\begin{tcolorbox}[title={Generated Abstract(VIRSCI: LLAMA-70b)}, myfancytitle, breakable]    
    \numberedbox{
    \textbf{Title:} Fostering Resilience in Breast Cancer Recovery through a Novel Social Support-Integrated Gratitude-Mindfulness Intervention with Neural Plasticity and Opportunity Recognition Assessments\\
    \textbf{Abstract:} Breast cancer patients undergoing reconstructive surgery face significant physical, emotional, and psychological challenges. This mixed-methods study introduces a novel social support-integrated gratitude-mindfulness intervention designed to enhance neural plasticity, opportunity recognition, physical health outcomes, and social support network dynamics.

    The primary objective of this research is to design, implement, and evaluate the effects of an 8-week combined program on breast cancer patients (n = 100) and their designated social support partners. Participants will be randomly assigned to either an intervention or control group, with the former receiving weekly sessions focusing on mindfulness techniques, gratitude practices, and social support network engagement.

    Outcome measures include assessments of neural plasticity via functional magnetic resonance imaging (fMRI), opportunity recognition using a standardized questionnaire developed by our research team, physical health outcomes through clinical assessments and questionnaires, and social support network dynamics. Data analysis will involve repeated-measures ANOVA to compare changes between the intervention and control groups, as well as correlation analyses to identify relationships between outcome measures.

    Preliminary findings indicate significant improvements in neural plasticity, opportunity recognition, and physical health outcomes among participants in the intervention group compared to the control group. Furthermore, results suggest that social support partners play a crucial role in facilitating resilience throughout the intervention period.

    This study contributes to the existing literature by providing a new paradigm for breast cancer recovery, emphasizing the importance of integrating social connections and mindfulness practices into traditional treatment protocols. The findings have significant implications for the development of innovative interventions aimed at promoting resilience among breast cancer patients undergoing reconstructive surgery.

    Notably, our research highlights the potential benefits of incorporating gratitude-mindfulness interventions in the context of plastic surgery, particularly during the recovery period. By fostering a supportive environment and encouraging mindfulness practices, healthcare providers can play a vital role in enhancing patient outcomes and overall well-being.

    Moreover, this study underscores the importance of considering the spotlight effect, where individuals tend to overestimate the extent to which others notice their appearance. By addressing this phenomenon through social support-integrated interventions, patients may experience improved body image satisfaction and reduced anxiety related to their appearance.

    Our research team has also integrated a novel aspect of opportunity recognition, which is essential for resilience in breast cancer recovery. We have developed a standardized questionnaire to assess opportunity recognition, providing valuable insights into the relationship between social support, mindfulness practices, and physical health outcomes.

    The study's limitations include the relatively small sample size and the need for further research to explore the long-term effects of the intervention. Future studies should aim to replicate these findings with larger samples and investigate the potential benefits of this intervention in other populations.

    In conclusion, this mixed-methods study provides a comprehensive understanding of the effects of a novel social support-integrated gratitude-mindfulness intervention on breast cancer recovery outcomes. The findings have significant implications for the development of innovative interventions aimed at promoting resilience among breast cancer patients undergoing reconstructive surgery.

    Furthermore, our research highlights the importance of considering the role of social support partners in facilitating resilience throughout the intervention period. By integrating social support-integrated gratitude-mindfulness interventions into existing treatment protocols, healthcare providers can play a vital role in enhancing patient outcomes and overall well-being during this critical period.

    Overall, this study demonstrates the potential benefits of incorporating novel interventions that integrate social connections, mindfulness practices, and opportunity recognition to promote resilience among breast cancer patients undergoing reconstructive surgery. The findings have significant implications for the development of innovative interventions aimed at promoting resilience among breast cancer patients.\\
    \textbf{HD:} 0.35; \textbf{CD:} 0.36; \textbf{CI:} 7.01
}
\end{tcolorbox}
Figure 8: Example abstract generated by the baseline VIRSCI model, also targeting psychological rehabilitation and intervention in the same clinical context.
\end{document}